\documentclass{article}
\usepackage{lmodern}

\usepackage[T1]{fontenc}
\usepackage[latin9]{inputenc}
\usepackage[english,american]{babel}
\usepackage{amsmath}
\usepackage{amsthm}
\usepackage{amssymb}
\usepackage[pdfusetitle,
 bookmarks=true,bookmarksnumbered=false,bookmarksopen=false,
 breaklinks=false,pdfborder={0 0 0},pdfborderstyle={},backref=false,colorlinks=false]
 {hyperref}
\hypersetup{
 pdfborderstyle=}

\makeatletter
\theoremstyle{plain}
\newtheorem{thm}{\protect\theoremname}[section]
\theoremstyle{definition}
\newtheorem{defn}[thm]{\protect\definitionname}
\theoremstyle{definition}
\newtheorem{example}[thm]{\protect\examplename}
\theoremstyle{remark}
\newtheorem{rem}[thm]{\protect\remarkname}

\usepackage{tikz}

\makeatother

\addto\captionsamerican{\renewcommand{\definitionname}{Definition}}
\addto\captionsamerican{\renewcommand{\examplename}{Example}}
\addto\captionsamerican{\renewcommand{\remarkname}{Remark}}
\addto\captionsamerican{\renewcommand{\theoremname}{Theorem}}
\addto\captionsenglish{\renewcommand{\definitionname}{Definition}}
\addto\captionsenglish{\renewcommand{\examplename}{Example}}
\addto\captionsenglish{\renewcommand{\remarkname}{Remark}}
\addto\captionsenglish{\renewcommand{\theoremname}{Theorem}}
\providecommand{\definitionname}{Definition}
\providecommand{\examplename}{Example}
\providecommand{\remarkname}{Remark}
\providecommand{\theoremname}{Theorem}

\begin{document}
\title{Batch normalization does not improve initialization}
\author{\selectlanguage{english}%
Joris Dannemann\thanks{Carl von Ossietzky Universität, Institut für Mathematik, 26129 Oldenburg,
Germany, E-mail: joris.dannemann@uni-oldenburg.de },\foreignlanguage{american}{ Gero Junike}\thanks{Corresponding author. Carl von Ossietzky Universität, Institut für
Mathematik, 26129 Oldenburg, Germany, ORCID: 0000-0001-8686-2661,
E-mail: gero.junike@uol.de}}
\maketitle
\selectlanguage{american}%
\begin{abstract}
Batch normalization is one of the most important regularization techniques
for neural networks, significantly improving training by centering
the layers of the neural network. There have been several attempts
to provide a theoretical justification for batch normalization. Santurkar
and Tsipras (2018) {[}How does batch normalization help optimization?
Advances in neural information processing systems, 31{]} claim that
batch normalization improves initialization. We provide a counterexample
showing that this claim is not true, i.e., batch normalization does
not improve initialization.\\
\textbf{Keywords:} neural network, batch normalization, initialization
\end{abstract}

\section{Introduction}

Neural networks are functions with many parameters. To train a neural
network, the parameters are randomly initialized at the beginning
of the training and are then calibrated to a learning set by minimizing
a loss function via gradient decent, see \cite{GoodBengCour16}. 

Batch normalization (BN), see \cite{ioffe2015batch}, is a technique
to improve the training of neural networks. It has been used in many
applications \cite{ioffe2015batch,GoodBengCour16,silver2017mastering,zhang2020matrix,flaig2022scenario,liang2025comprehensive}
to make training faster and more stable. According to \cite[p. 317]{GoodBengCour16},
``Batch normalization (Ioffe and Szegedy, 2015) is one of the most
exciting recent innovations in optimizing deep neural networks''.
Similarly, \cite[p. 1]{bjorck2018understanding} concludes: ``Nowadays,
there is little disagreement in the machine learning community that
BN accelerates training, enables higher learning rates, and improves
generalization accuracy {[}...{]} and BN has successfully proliferated
throughout all areas of deep learning''. 

However, the working mechanism of BN is not yet fully understood.
The authors in \cite{santurkar2018does} are among the first to analyze
BN from a more theoretical point of view. They provide several theoretical
arguments why BN improves training, in particular they claim in Lemma
4.5 ``that BN also offers an advantage in initialization'', see
\cite[p. 8]{santurkar2018does}. This statement, i.e., \cite[Lemma 4.5]{santurkar2018does},
is also cited in \cite[ Lemma 7.2]{zhang2020matrix}. However, we
provide a straightforward counterexample showing that this lemma is
incorrect, i.e., BN does not improve initialization. 

This article is structured as follows: in Section \ref{sec:Neural-Networks-and}
we give a brief introduction to neural networks and BN. In Section
\ref{sec:Counterexample:-BN-does}, we provide a counterexample showing
that BN does not improve initialization.

\section{\protect\label{sec:Neural-Networks-and}Neural networks and batch
normalization}

For an introduction to neural networks with BN, see \cite[Section 8.7.1]{GoodBengCour16}
and \cite[Section 7.15]{zhang2020matrix}. In this section, we summarize
the most important definitions: A \emph{neural network} is a function
from some \emph{input space} (e.g., a subspace of $\mathbb{R}^{p}$)
to some \emph{output space} (e.g., a subspace of $\mathbb{R}$). A
neural network consists of\emph{ layers,} that consist of \emph{neurons},
which use \emph{activation functions}. The regularization technique
\emph{batch normalization} acts on one or more layers. We write $\Vert\cdot\Vert$
and $\langle\cdot,\cdot\rangle$ for the Euclidean norm and the inner
product on $\mathbb{R}^{d}$, respectively.
\begin{defn}
An \emph{activation function} is a function $g:\mathbb{R}\to\mathbb{R}$. 
\end{defn}

For example, the identity, Sigmoid, tanh and ReLu (Rectified linear
unit) activation functions are defined by $x\mapsto x$, $x\mapsto\frac{1}{1+e^{-x}}$,
$x\mapsto\tanh(x)$ and $x\mapsto\max(x,0)$, respectively. Next,
we define a neuron, that has $k$ inputs and one output. 
\begin{defn}
A \emph{neuron} is a function $h$ with\emph{ weights} $\boldsymbol{w}\in\mathbb{R}^{k}$
and \emph{bias} $b\in\mathbb{R}$ and activation function $g$, such
that 
\begin{align*}
h: & \mathbb{R}^{k}\to\mathbb{R},\quad\boldsymbol{x}\mapsto g\big(\langle\boldsymbol{w},\boldsymbol{x}\rangle+b\big).
\end{align*}
\end{defn}

\begin{defn}
\label{defn1}Let $h_{1},...,h_{n}:\mathbb{R}^{k}\to\mathbb{R}$ be
$n$ neurons. A layer is a function
\[
f:\,\mathbb{R}^{k}\to\mathbb{R}^{n},\quad\boldsymbol{x}\mapsto\big(h_{1}(\boldsymbol{x}),...,h_{n}(\boldsymbol{x})\big)^{\intercal}
\]
consisting of $n\in\mathbb{N}$ neurons. Each neuron has $k$ inputs. 
\end{defn}

\begin{example}
In the special case that each neuron of the layer has the identity
as its activation function and zero bias, it holds that $f(\boldsymbol{x})=W\boldsymbol{x}$
for a matrix $W\in\mathbb{R}^{n\times k}$ of weights. 
\end{example}

\begin{defn}
\label{defn2}Suppose the input space has dimension $p$. A \emph{neural
network} (NN)\emph{ }consists of $L\in\mathbb{N}$ layers $f_{1}:\mathbb{R}^{k_{1}}\to\mathbb{R}^{n_{1}}$,...,$f_{L}:\mathbb{R}^{k_{L}}\to\mathbb{R}^{n_{L}}$
with $k_{1}=p$, $k_{2}=n_{1}$,..., $k_{L}=n_{L-1}$. The layers
have $n_{1},n_{2},...,n_{L}$ neurons. The neural network is the function
\[
\varphi_{\Theta}:\mathbb{R}^{p}\to\mathbb{R}^{n_{L}},\quad\boldsymbol{x}\mapsto f_{L}\big(f_{L-1}\big(...f_{2}\big(f_{1}(\boldsymbol{x})\big).
\]
The input $\boldsymbol{x}$ is also called the \emph{input layer}.
The last layer $f_{L}$ is also called the \emph{output layer}. The
layers $f_{1},...,f_{L-1}$ are called \emph{hidden layers.} All parameters
of the NN, e.g., the weights and the biases of the neurons, are stored
in a vector $\Theta$.
\end{defn}

The optimal parameters $\Theta^{\ast}$ of the NN are usually found
via minimizing some loss function, e.g.,
\[
\Theta^{\ast}=\text{argmin}_{\Theta}\frac{1}{N}\sum_{i=1}^{N}\bigg(\varphi_{\Theta}\big(\bar{\boldsymbol{x}}^{(i)}\big)-\bar{y}^{(i)}\bigg)^{2},
\]
where $\bar{\boldsymbol{x}}^{(i)}\in\mathbb{R}^{p}$ and $\bar{y}^{(i)}\in\mathbb{R}$
and $\big\{\big(\bar{\boldsymbol{x}}^{(1)},\bar{y}^{(1)}\big),...,\big(\bar{\boldsymbol{x}}^{(N)},\bar{y}^{(N)}\big)\big\}$
is a learning set of size $N$. The loss function is minimized by
(stochastic) gradient descent. It is recommended to initialize $\Theta$
randomly, see \cite{glorot2010understanding,he2015delving}. During
the optimization, the learning set is divided into several disjoint
sets, so called \emph{batches}.

Next, we define layers with BN, introduced by \cite{ioffe2015batch}.
Note that the input layer of a batch of size $M$ has dimension $p\times M$
and the output of the $l$-th layer has dimension $n_{l}\times M$.
\begin{defn}
\label{defn3}Consider a NN with of $L$ layers. The $l$-th layer
$f_{l}:\mathbb{R}^{k_{l}}\to\mathbb{R}^{n_{l}}$ consists of $n_{l}$
neurons. Apply the NN to a batch of size $M\in\mathbb{N}$. The output
of the $l$-th layer has the form
\[
\mathbb{\mathbb{Z}}_{l}=\left(\begin{array}{ccc}
z_{11} & ... & z_{1M}\\
\vdots &  & \vdots\\
z_{n_{l}1} & ... & z_{n_{l}M}
\end{array}\right)\in\mathbb{R}^{n_{l}\times M}.
\]
Let
\[
\mu_{j}=\frac{1}{M}\sum_{m=1}^{M}z_{jm},\quad\text{and \ensuremath{\quad\sigma_{j}=\sqrt{\frac{1}{M}\sum_{m=1}^{M}(z_{jm}-\mu_{j})^{2}}}},\quad j=1,...,n_{l}.
\]
Let $\gamma_{1},...,\gamma_{n_{l}}\in\mathbb{R}$ and $\beta_{1},...,\beta_{n_{l}}\in\mathbb{R}$
be parameters and define
\[
\hat{\mathbb{\mathbb{Z}}}_{l}=\left(\begin{array}{ccc}
\gamma_{1}\frac{z_{11}-\mu_{1}}{\sigma_{1}}+\beta_{1} & ... & \gamma_{1}\frac{z_{1M}-\mu_{1}}{\sigma_{1}}+\beta_{1}\\
\vdots &  & \vdots\\
\gamma_{n_{l}}\frac{z_{n_{l}1}-\mu_{n_{l}}}{\sigma_{n_{l}}}+\beta_{n_{l}} & ... & \gamma_{n_{l}}\frac{z_{n_{l}M}-\mu_{n_{l}}}{\sigma_{n_{l}}}+\beta_{n_{l}}
\end{array}\right)\in\mathbb{R}^{n_{l}\times M}.
\]
$\hat{\mathbb{\mathbb{Z}}}_{l}$ is called\emph{ batch normalization
transform }of layer\emph{ $\mathbb{\mathbb{Z}}_{l}$.}
\end{defn}

\begin{rem}
During the training, the NN is trained on the transformed data $\hat{\mathbb{\mathbb{Z}}}_{l}$
and the parameters $\gamma_{j}$ and $\beta_{j}$, $j=1,...,n_{l}$
are trainable as well. If $\gamma_{j}=\sigma_{j}$ and $\beta_{j}=\frac{\mu_{j}}{\sigma_{j}}$
then $\hat{\mathbb{\mathbb{Z}}}_{l}=\mathbb{Z}_{l}$, i.e., one ``could
recover the original activations, if that were the optimal thing to
do'', see \cite[Section 3, p. 3]{ioffe2015batch}. For prediction,
the average values (over all batches) of the parameters $\mu_{j}$
and $\sigma_{j}$ are used.
\end{rem}

\section{\protect\label{sec:Counterexample:-BN-does}Counterexample: BN does
not improve initialization}

\cite{santurkar2018does} analyze individual layers of a NN assuming
that all neurons in that layer have the identity as activation function
and that $\beta$ and $\gamma$ for the BN are constants. In \cite[Lemma 4.5]{santurkar2018does}
it is claimed that BN offers an advantage in initialization:

``(BatchNorm leads to a favorable initialization). \emph{Let $\boldsymbol{W}^{\boldsymbol{\ast}}$
and $\widehat{\boldsymbol{W}}^{\boldsymbol{*}}$ be the set of local
optima for the weights in the normal and BN networks, respectively.
For any initialization $W_{0}$ 
\begin{equation}
\left\Vert W_{0}-\widehat{W}^{*}\right\Vert ^{2}\leq\Vert W_{0}-W^{*}\Vert^{2}-\frac{1}{\Vert W^{*}\Vert^{2}}\left(\Vert W^{*}\Vert^{2}-\langle W^{*},W_{0}\rangle\right)^{2},\label{eq:W0Wstar}
\end{equation}
if $\langle W_{0},W^{*}\rangle>0$, where $\widehat{W}^{*}$ and $W^{*}$
are closest optima for BN and standard network, respectively.''},
see \cite[p. 9]{santurkar2018does}. 
\begin{rem}
The matrix $W_{0}\in\mathbb{R}^{n\times k}$ is understood as the
initial weights of a particular layer with $k$ inputs consisting
of $n$ neurons with biases equal to zero. $W^{*}$ and $\widehat{W}^{*}$
are the locally optimal weights of the layer with respect to the loss
function with and without BN, respectively. The norm and the inner
product applied to matrices in Inequality (\ref{eq:W0Wstar}) are
understood column wise. 
\end{rem}

In the next example, we will show that \cite[Lemma 4.5]{santurkar2018does}
is wrong. We will construct a NN consisting of a single neuron with
the identity as activation function with initial weights $W_{0}$
identical to the optimal weights $W^{\ast}$ and show that any weights
$\widehat{W}^{*}$ satisfying Inequality (\ref{eq:W0Wstar}) cannot
be optimal for the BN network.
\begin{example}
\label{exa:We-consider-a}We consider a learning set $\big\{\big(\bar{\boldsymbol{x}}^{(1)},\bar{y}^{(1)}\big),...,\big(\bar{\boldsymbol{x}}^{(3)},\bar{y}^{(3)}\big)\big\}$
consisting of three samples, where 
\[
\bar{\boldsymbol{x}}^{(1)}=\begin{pmatrix}1\\
1
\end{pmatrix},\quad\bar{\boldsymbol{x}}^{(2)}=\begin{pmatrix}1\\
2
\end{pmatrix},\quad\bar{\boldsymbol{x}}^{(3)}=\begin{pmatrix}2\\
3
\end{pmatrix},\quad\bar{y}^{(1)}=2,\quad\bar{y}^{(1)}=5,\quad\bar{y}^{(1)}=13.
\]
There is the following relation: $\bar{y}^{(i)}=\big(\bar{x}_{1}^{(i)}\big)^{2}+\big(\bar{x}_{2}^{(i)}\big)^{2}$.
To learn this relation, we consider a NN consisting of a single neuron
and the identity as activation function: $\varphi_{\boldsymbol{w}}(\boldsymbol{x})=\left\langle \boldsymbol{w},\boldsymbol{x}\right\rangle $,
where $\boldsymbol{w}=(w_{1},w_{2})\in\mathbb{R}^{2}$ are the weights
of the neuron. The bias is set to zero. We initialize the NN with
the weights $W_{0}:=(1,3)$. We consider a batch identical to the
learning set. The output matrix of the hidden layer of the NN for
the batch is therefore (we omit the index $l$ since there is only
one layer):
\[
\mathbb{\mathbb{Z}}=\begin{pmatrix}w_{1}+w_{2} & w_{1}+2w_{2} & 2w_{1}+3w_{2}\end{pmatrix}\in\mathbb{R}^{1\times3}.
\]
The mean $\mu$ and standard-deviation $\sigma$ over the batch are
then given by (the index $j$ omitted)
\[
\mu=\frac{4}{3}w_{1}+2w_{2}
\]
and 
\[
\sigma=\frac{1}{3}\sqrt{2w_{1}^{2}+6w_{1}w_{2}+6w_{2}^{2}}.
\]
We choose $\gamma=1$ and $\beta=0$ as parameters for the BN transform.
This means the NN with BN is given by
\[
\widehat{\varphi}_{\boldsymbol{w}}(\boldsymbol{x})=\frac{\left\langle \boldsymbol{w},\boldsymbol{x}\right\rangle -\mu}{\sigma}.
\]
For the cost function we use the squared Euclidean norm. More specifically,
for the standard network, let 
\begin{align*}
\mathcal{C}(\boldsymbol{w})= & \sum_{i=1}^{3}\left(\varphi_{\boldsymbol{w}}\big(\bar{\boldsymbol{x}}^{(i)}\big)-\bar{y}^{(i)}\right)^{2}\\
= & (w_{1}+w_{2}-2)^{2}+(w_{1}+2w_{2}-5)^{2}+(2w_{1}+3w_{2}-13)^{2}.
\end{align*}
This function achieves its global minimum at $W^{*}=(1,3)$. The cost
function for the BN network $\widehat{\varphi}_{\boldsymbol{w}}$
is given by
\begin{align*}
\widehat{\mathcal{C}}(\boldsymbol{w})=\sum_{i=1}^{3}\left(\widehat{\varphi}_{\boldsymbol{w}}\big(\bar{\boldsymbol{x}}^{(i)}\big)-\bar{y}^{(i)}\right)^{2}= & \left(\frac{-w_{1}-3w_{2}}{\sqrt{2w_{1}^{2}+6w_{1}w_{2}+6w_{2}^{2}}}-2\right)^{2}\\
+ & \left(\frac{-w_{1}}{\sqrt{2w_{1}^{2}+6w_{1}w_{2}+6w_{2}^{2}}}-5\right)^{2}\\
+ & \left(\frac{2w_{1}+3w_{2}}{\sqrt{2w_{1}^{2}+6w_{1}w_{2}+6w_{2}^{2}}}-13\right)^{2}.
\end{align*}
The gradient of $\widehat{\mathcal{C}}$ at $W^{*}$ does not vanish,
it is given by $(-0.34,0.11)$. In particular, $\widehat{\mathcal{C}}$
cannot attain a local minium at $W^{*}$. Since $W_{0}=W^{*}$, it
holds that
\[
\langle W_{0},W^{*}\rangle=\langle W_{0},W_{0}\rangle=\Vert W_{0}\Vert^{2}>0.
\]
It can be shown that $\widehat{\mathcal{C}}$ achieves a local minimum
at any $(w_{1},w_{2})$ satisfying $w_{1}>0$ and $3w_{1}=5w_{2}$,
which is a consequence of BN being scale invariant. Thus, $\widehat{\boldsymbol{W}}^{\boldsymbol{*}}\neq\emptyset$.
Let $\widehat{W}^{*}\in\widehat{\boldsymbol{W}}^{\boldsymbol{*}}$.
Assume the statement in \cite[Lemma 4.5]{santurkar2018does} were
true. Then by Inequality (\ref{eq:W0Wstar}) it holds that
\[
\Vert W_{0}-\widehat{W}^{*}\Vert^{2}\leq\underbrace{\Vert W_{0}-W^{*}\Vert^{2}}_{=0}-\frac{1}{\Vert W^{*}\Vert^{2}}\big(\underbrace{\Vert W^{*}\Vert^{2}-\langle W^{*},W_{0}\rangle}_{=\Vert W^{*}\Vert^{2}-\Vert W^{*}\Vert^{2}=0}\big)^{2}=0,
\]
which implies $\widehat{W}^{*}=W_{0}$. Hence, $W_{0}$ would also
be a local optimum for the cost function of the BN network. However,
this does not hold since the gradient of $\widehat{\mathcal{C}}$
does not vanish at $W_{0}$. Hence, the statement in \cite[Lemma 4.5]{santurkar2018does}
is wrong. In conclusion, BN does not improve initialization.
\end{example}

\begin{rem}
The (incorrect) proof of this statement can be found in \cite{santurkar2018doesArxiv}.
Let $W$ be some weights of a layer. In the proof it is claimed that\textit{\emph{
if $W$ leads to a local optimum of the loss function before BN then
$W$ also leads to a local optimum after BN, which is not correct:
The cost function of the standard network in Example \ref{exa:We-consider-a}
has a local minimum at $W^{*}=(1,3)$ but the cost function $\widehat{\mathcal{C}}$
of the BN network cannot have a local minimum at $W^{\ast}$ since
the gradient of $\widehat{\mathcal{C}}$ does not vanish at $W^{\ast}$. }}
\end{rem}

\subsection*{Disclosure statement. }

The authors report there are no competing interests to declare. 

\bibliographystyle{plain}
\bibliography{biblio}

\end{document}